\documentclass[conference]{IEEEtran}
\IEEEoverridecommandlockouts
\usepackage{cite}
\usepackage{amsmath,amssymb,amsfonts}
\usepackage{algorithmic}
\usepackage{graphicx}
\usepackage{textcomp}
\usepackage{xcolor}
\usepackage{booktabs}
\usepackage{listings}
\usepackage{caption}
\usepackage{multirow}
\usepackage{tablefootnote}
\usepackage[newfloat,frozencache,cachedir=minted-cache]{minted}

\def\BibTeX{{\rm B\kern-.05em{\sc i\kern-.025em b}\kern-.08em
    T\kern-.1667em\lower.7ex\hbox{E}\kern-.125emX}}

\usepackage{tikz}
\usepackage{textcomp}
\usepackage{hyperref}
\usepackage{lipsum}

\newcommand\copyrighttext{%
  \footnotesize \textcopyright 2023 IEEE. Personal use of this material is permitted.
  Permission from IEEE must be obtained for all other uses, in any current or future
  media, including reprinting/republishing this material for advertising or promotional
  purposes, creating new collective works, for resale or redistribution to servers or
  lists, or reuse of any copyrighted component of this work in other works.}
\newcommand\copyrightnotice{%
\begin{tikzpicture}[remember picture,overlay]
\node[anchor=south,yshift=10pt] at (current page.south) {\fbox{\parbox{\dimexpr\textwidth-\fboxsep-\fboxrule\relax}{\copyrighttext}}};
\end{tikzpicture}%
}

\begin{document}

 \title{Open the box of digital neuromorphic processor: Towards effective algorithm-hardware co-design}
%\title{Open the Box of Neuromorphic Processors For Effective Algorithm-Hardware Benchmark and Co-Design}

\author{\IEEEauthorblockN{Guangzhi Tang\IEEEauthorrefmark{1}, Ali Safa\IEEEauthorrefmark{2}\IEEEauthorrefmark{3}, Kevin Shidqi\IEEEauthorrefmark{1}, Paul Detterer\IEEEauthorrefmark{1}, Stefano Traferro\IEEEauthorrefmark{1},}
\IEEEauthorblockN{Mario Konijnenburg\IEEEauthorrefmark{1}, Manolis Sifalakis\IEEEauthorrefmark{1}, Gert-Jan van Schaik\IEEEauthorrefmark{1}, Amirreza Yousefzadeh\IEEEauthorrefmark{1}}
\IEEEauthorblockA{\textit{\IEEEauthorrefmark{1}imec Netherlands, Eindhoven, Netherlands}, \textit{\IEEEauthorrefmark{2}imec, Leuven, Belgium}, \textit{\IEEEauthorrefmark{3}KU Leuven, Leuven, Belgium}}
}

\maketitle
\copyrightnotice

\begin{abstract}
Sparse and event-driven spiking neural network (SNN) algorithms are the ideal candidate solution for energy-efficient edge computing. Yet, with the growing complexity of SNN algorithms, it isn't easy to properly benchmark and optimize their computational cost without hardware in the loop. Although digital neuromorphic processors have been widely adopted to benchmark SNN algorithms, their black-box nature is problematic for algorithm-hardware co-optimization. In this work, we open the black box of the digital neuromorphic processor for algorithm designers by presenting the neuron processing instruction set and detailed energy consumption of the SENeCA neuromorphic architecture. For convenient benchmarking and optimization, we provide the energy cost of the essential neuromorphic components in SENeCA, including neuron models and learning rules. Moreover, we exploit the SENeCA's hierarchical memory and exhibit an advantage over existing neuromorphic processors. We show the energy efficiency of SNN algorithms for video processing and online learning, and demonstrate the potential of our work for optimizing algorithm designs. Overall, we present a practical approach to enable algorithm designers to accurately benchmark SNN algorithms and pave the way towards effective algorithm-hardware co-design.
\end{abstract}

\section{Introduction}

Energy-efficient computations are essential for edge applications that operate with limited energy resources. Brain-inspired spiking neural networks (SNNs) have the potential to reduce energy costs by introducing sparse and event-driven computation~\cite{maass1997networks}, making them ideal candidate solutions for the edge. However, the low-power assumption of the SNN algorithms is not always valid if computational costs are not properly benchmarked. Many works use the sparsity of synaptic operations to demonstrate efficiency~\cite{sengupta2019going, kim2020spiking,cordone2021learning}, disregarding additional expenses introduced by hardware primitives like memory access or instruction operation. Since SNN algorithms require dedicated hardware, namely the neuromorphic processor, algorithm designs based on inaccurate hardware assumptions can fail to realize potential advantages. Therefore, there is a need for effective algorithm-hardware co-design to truly realize the promised benefits of neuromorphic computing.

Digital neuromorphic processors provide the opportunity to benchmark the energy efficiency of SNNs~\cite{merolla2014million,davies2018loihi,moreira2020neuronflow,kumar2022decoding,tang2020reinforcement}. However, these processors behave like a black box for algorithm designers. First, their bottom-up designs support restricted predefined computational elements and leave limited space for co-optimizing new algorithms with the hardware. Second, the coarse benchmarking results from the hardware do not provide precise insight into the design of the SNN algorithm to locate potential optimizations. Although there are neuromorphic processors developed using co-design approaches~\cite{frenkel202028,fang2021neuromorphic,zhong2021spike, datta2022ace}, they are mainly confined to a specific SNN algorithm and are hard to use by algorithm designers without a sufficient hardware background. Therefore, algorithm designers need a flexible neuromorphic processor design with transparent and customizable internal operations.

\begin{figure}[t]
\centering
\includegraphics[scale=0.45]{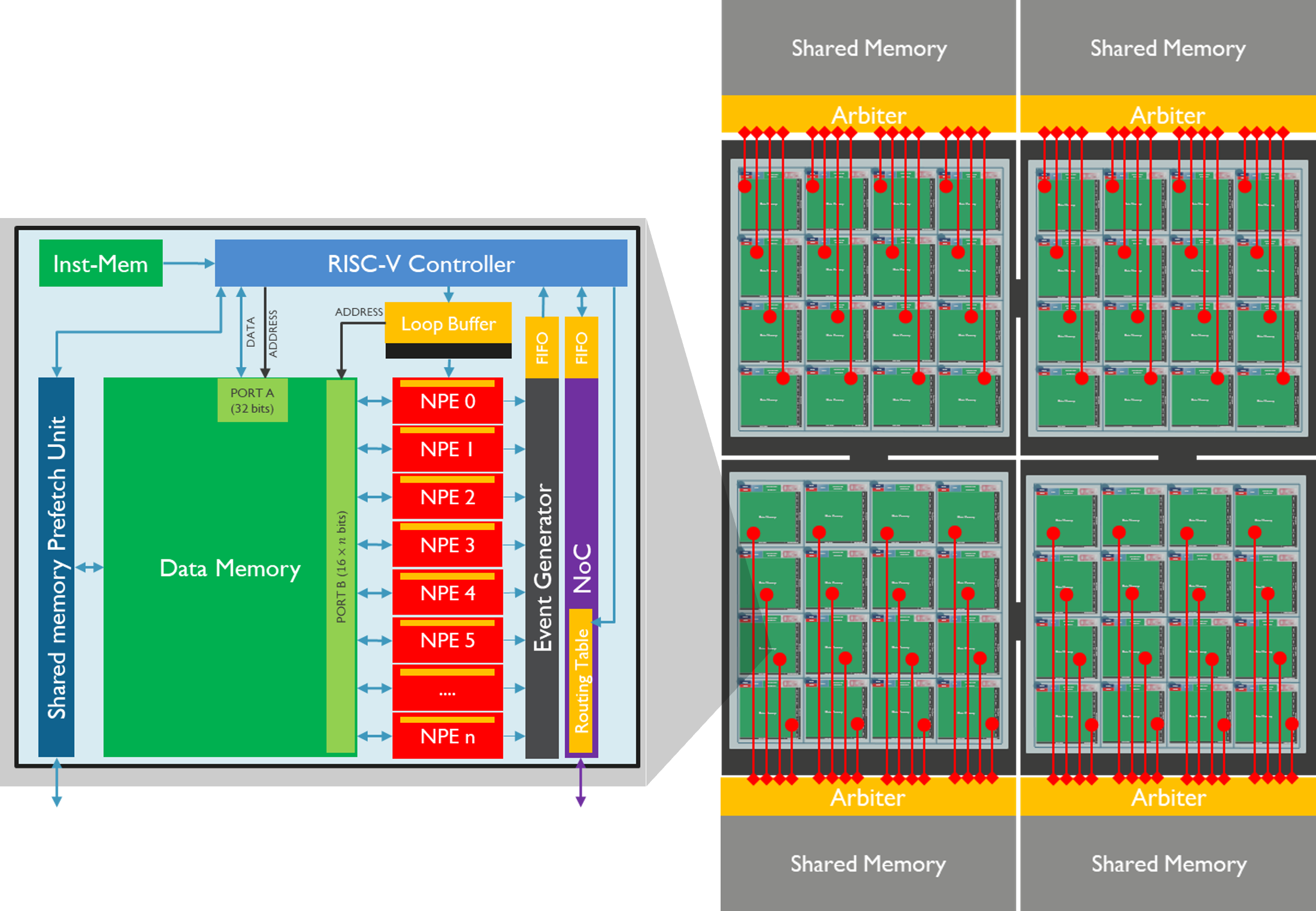}
\caption{The pipeline of a SENeCA neuromorphic core (left), the interconnected mesh architecture via NoC (right), and the hierarchical memory consist of register files (orange), local SRAM memories (green) and large shared memories (gray).}
\label{fig:seneca}
\end{figure}

In this work, we precisely detail the neuron processing instruction set of SENeCA~\cite{yousefzadeh2022seneca}, our scalable and flexible digital neuromorphic architecture, to help algorithm designers conveniently benchmark and optimize the cost of their novel SNN algorithms. To demonstrate the potential of SENeCA on algorithm-hardware co-design, we show three levels of abstraction to benchmark costs for SNN algorithms. The main contributions of this paper are the following: 
\begin{enumerate}
    \item We conduct circuit-level energy measurements on neuron processing instructions in SENeCA (Section \ref{sec1}). This will enable the algorithm design community to accurately estimate the energy cost of their novel SNN algorithms without running them on the actual hardware.
    \item We provide essential neuromorphic components (neuron models, learning rules, and hierarchical memory exploitation) constructed using SENeCA instructions, together with their energy costs (Section \ref{microkernels}). This layer of abstraction will enable algorithm designers to quickly estimate hardware overheads of typical SNN topologies without resorting to low-level instruction.
    \item To clearly verify the usefulness of our contributions, we illustrate how our framework can be utilized to compute the energy efficiency of different SNN algorithms targeting video processing and online learning (Section \ref{applevel}), based on the energy costs provided in this work.
\end{enumerate}

%This paper is organized as follows. The SENeCA architecture, together with the energy measurements of each neuron processing instruction are presented in Section \ref{sec1}. A number of typical neuromorphic components, together with their micro-code implementation and energy consumption are provided in Section \ref{microkernels}. Then, Section \ref{applevel} illustrates how our framework can be used to estimate the energy cost of a different SNN algorithm. Finally, conclusions are provided in Section \ref{conc}.
%First, we measured the energy consumption of neuron processing instructions to serve as a basis for accurate cost estimation of SNN algorithms without running them on the hardware. Second, we provide essential neuromorphic components constructed from the instructions, including neuron models, learning rules, and methods to exploit hierarchical memory, to add a layer of abstraction for easier algorithm optimization. Third, we compute the energy efficiency of SNN algorithms for video processing and online learning based on the cost of neuromorphic components.

\section{Neuron Processing on Neuromorphic Processor}
\label{sec1}
The SENeCA neuromorphic architecture performs event-driven computation with time-multiplexing Neuron Processing Elements (NPEs) emulating numerous neurons per core (Figure \ref{fig:seneca}). To provide sufficient flexibility, SENeCA embeds a RISC-V controller that enables customizable processing pipelines, rich NPE instructions for versatile computations, and hierarchical memories to optimize the deployment and processing of networks. When a new event enters the core, the RISC-V is interrupted from sleep, preprocesses the event, writes information into the NPEs, and activates the neuron processing before returning to sleep. After events are captured from NPEs, they interrupt the RISC-V from sleep again and communicate to other cores via the NoC.

\vspace{-2pt}
\subsection{NPE and Neuron Processing Instruction Set}

NPE is the central neuron processing unit in the SENeCA core, which accelerates a rich neuron processing instruction set (Table \ref{tableinst}). Each instruction is executed in one cycle (pipelined, 2ns per cycle) and operates in BrainFloat 16 (BF16) format~\cite{kalamkar2019study}. SNN algorithms can be built by different sequential executions of the instructions, namely micro-kernels. These micro-kernels are stored in the register-files of loop buffer and sent to the NPEs during runtime. For efficient time-multiplexing, the loop buffer executes micro-kernels in a "for-loop" fashion on NPEs and incrementally calculates Data-Memory addresses. This design gives a much lower cost than using the more flexible instruction memory (Table \ref{tablemem}). Determined by the event type, the RISC-V controller selects which micro-kernel to process on the NPEs. Neuron processing operates with hierarchical memory, including register-files, local data memory, and external shared memory if the model cannot fit locally. To introduce intra-core parallelism, NPEs in the SENeCA core form a SIMD (single instruction multiple data) type architecture~\cite{flynn1972some} that accesses data through a wide data memory port in parallel. The NPE also supports quantized integer data types (Int4 and Int8) to reduce energy costs (see Section \ref{intweight}). When events are generated, the event capture unit converts them to the Address Event Representation (AER) form \cite{6865445} before sending them to the RISC-V and NoC. The present version of SENeCA core has 8 NPEs and 64 registers per NPE. These numbers are parameterized and can be fine-tuned before synthesis.

\vspace{-2pt}
\subsection{Circuit-level Energy Measurements}

We report the average consumption of the NPE instructions in Table \ref{tableinst}. The pre-silicon energy number includes the power consumption of all the modules needed to execute the instruction (e.g., address calculations in loop buffer, access to instruction memory, etc.). The results are measured by running each instruction 8k times with random data using the Cadence JOULES (time-based mode), a RTL level power measurement tool (within 15\% of signoff power)\cite{JOULES}, with the GF-22nm FDX technology node\footnote{In typical corner (0.8v and 25C, no back-biasing) }. The leakage power for the core is around 30$\mu$W (0.06pJ in a 2ns clock cycle). For clarity, we report the memory and NoC information in Table \ref{tablemem}. Since in a typical SNN, there are significantly more synaptic operations than events, the computational cost for synaptic operations (done in NPEs) largely dominates the event pre-processing (RISC-V) and communications (NoC). Therefore, in this paper, for simplicity and due to limited space, we safely ignored the RISC-V and NoC costs.

\section{Essential Neuromorphic Components}
\label{microkernels}
Direct optimization of complex algorithms at the instruction level is difficult. A level of abstraction for essential components of the SNN algorithm can significantly simplify benchmarking and optimization. Here, we present neuron models and learning rules constructed from NPE instructions and compute their cost using circuit-level power measurements (see Table \ref{tablecomponents}). Furthermore, we exploit the hierarchical memory in SENeCA and compare the costs of synaptic operations when using quantized integer weights and multi-event processing.

\begin{table}[t]
\vspace{5.2pt}
\small
  \caption{Neuron Processing Energy Consumption}
  \label{tableinst}
  \centering
  \begin{tabular}{|c|c|c|}
    \hline
    Instruction & Description & Energy (pJ) \\
    \hline
    ADD/SUB/MUL/DIV & Arithmetic ops. & 1.4 \\
     & 2xINT8b Arithmetic ops. & 1.2\\
    %INT ADD(8bit+4bit) &              & 0.6 \\
    \hline
    GTH/MAX/MIN    & Compare ops.  & 1.2 \\
    EQL/ABS        &               & 1.1 \\
    \hline
    AND/ORR & Bit-wise ops. & 1.1 \\
    SHL/SHR        &               & 1.2 \\
    \hline
    % I2F(Rdst, R1, Rf) & $Rdst=BF16(R1)$ & 0.62\\
    I2F & data type cnv. & 1.1\\
    RND        &               & 1.4 \\
    \hline
    % EVC(r, tag, inc) & Event Capture & 0.21\\
    %                  & + if generates event & + 0.15\\
    EVC  & Event Capture & 0.5\\
                     & + if generates event & + 1.1\\
    \hline
    % MLD(Rdst, Raddr, inc) & Read Data Mem & 3.6\\
    % MST(Raddr, Rdst, inc) & Write Data Mem & 3.6\\
    MLD & Data Mem Load/Store  & 3.7 \\
    MST        &               & 3.9 \\
    \hline
    RISC-V & Per Instruction & 11.6\\
    pre/post Process & + Data mem access & +10.0\\
    % NoC Process & Single Event & 0.0\\
    \hline
  \end{tabular}
\end{table}

% \begin{table}[t]
% \small
%   \caption{Memory Size and Energy Consumption}
%   \label{tablemem}
%   \centering
%   \begin{tabular}{|l|c|c|}
%     \hline
%      &  NPE Registers & Data Mem (SRAM)\\
%     \hline
%     Size & $64W\times 16b$ & $8KW\times 32b$ ($2Mb$)\\
%     \hline
%     % Energy (pJ) & 0.14 & 25.39\\
%     % Energy (fJ/b) & 8.75 & 198.36\\
%     Energy (fJ/b) & 12.0 & ?\\
%     \hline
%     \hline
%     & Loop Buffer & Inst Mem (SRAM)\\
%     \hline
%     Size & $128W\times 17b$ & $8KW\times 32b$ ($128Kb$)\\
%     \hline
%     Energy (fJ/b) & 35.29 & 179.68\\
%     \hline
%     \hline
%     & NoC event  & Shared Mem (HBM)~\cite{HBM}
%     %\footnote{number is based on the reported power consumption in https://www.xilinx.com/products/silicon-devices/fpga/virtex-ultrascale-plus-hbm.html}
%     \\
%     \hline
%     Size & $32b$ & $32b$ (multi Gb)\\
%     \hline
%     % Energy (pJ) & 2.1 & 224\\
%     Energy (fJ/b) & 65.62 & 7000\\
%     \hline
%   \end{tabular}
% \vspace{-10pt}
% \end{table}

\begin{table}[t]
\small
  \caption{Memory Size and Energy Consumption}
  \label{tablemem}
  \centering
  \begin{tabular}{|l|c|c|}
    \hline
     &  Register-File (NPE) & SRAM (Inst/Data Mem)\\
    \hline
    Size & $64W\times 16b$ & $8KW\times 32b$ ($2Mb$)\\
    \hline
    % Energy (pJ) & 0.14 & 25.39\\
    % Energy (fJ/b) & 8.75 & 198.36\\
    Energy (fJ/b) & 12.0 & 200\\
    \hline
    \hline
    & NoC event  & HBM (Shared Mem)~\cite{HBM}
    %\footnote{number is based on the reported power consumption in https://www.xilinx.com/products/silicon-devices/fpga/virtex-ultrascale-plus-hbm.html}
    \\
    \hline
    Size & $32b$ & $32b$ (multi Gb)\\
    \hline
    % Energy (pJ) & 2.1 & 224\\
    Energy (fJ/b) & 65.62 & 7000\\
    \hline
  \end{tabular}
\vspace{-10pt}
\end{table}

\subsection{Integrate and Fire Neuron}
\label{IFneuron}
Integrate and Fire (IF) neurons are widely used for SNN processing~\cite{abrahamsen2004time,indiveri2010spike,stuijt2021mubrain}. Here, we define an IF neuron as:
\begin{equation}
\small
\begin{gathered}
    v_{i}[k] \xleftarrow{} v_{i}[k-1] \times (1 - s_{out,i}[k-1]) + \Sigma_j w_{ij}\times s_{in,j}[k] \\
    s_{out,i}[k] \xleftarrow{} H(v_{i}[k]-v_{th})
\end{gathered}
\label{ifneuron}
\end{equation}
where $k$ is the time step, $v_i$ is the state of neuron $i$, $s_{in,j}$ is the input spike from neuron $j$, $s_{out,i}$ is the output spike of neuron $i$, $w_{ij}$ is the weight, $v_{th}$ is the voltage threshold and $H$ is the Heaviside function. The first micro-kernel in Component \ref{microifneuron} integrates spikes instantly, and the second micro-kernel generates spikes at the end of each time step.

\subsection{Sigma Delta Neuron}

Sigma Delta (SD) neurons sparsify deep neural networks (DNNs) by communicating temporal activation differences through events~\cite{o2016sigma}. First, the sigma integrates events as:
\begin{equation}
\small
    z_i[k] \xleftarrow{} z_i[k-1] + \Sigma_j w_{ij}\times o_{in, j}[k]
\label{sigma}
\end{equation}
where $z_i$ is the sigma state of neuron $i$ and $o_{in, j}$ is the input event from neuron $j$. Then, the delta generates events as:
\begin{equation}
\small
    o_{out, i}[k] \xleftarrow{} round(\frac{f(z_i[k])}{q})\times q - round(\frac{f(z_i[k-1])}{q})\times q
\label{delta}
\end{equation}
where $o_{out, i}$ is the output event of neuron $i$, $f$ is a non-linear activation function (e.g. $ReLU$), $round$ function rounds a number to integer and $q$ is the scaling factor. The quantization can significantly increase the sparsity of the events~\cite{yousefzadeh2022delta}. The first micro-kernel in Component \ref{microsdneuron} integrates events instantly and the second operates at flexible frequency while maintaining equivalence to the trained DNN~\cite{yousefzadeh2019asynchronous}.

\begin{listing}[t]
\vspace{5.2pt}
\small Micro-kernel 1: \textit{\textbf{Spike Integration}. See Eq. (\ref{ifneuron}).}
\begin{minted}[frame=single,escapeinside=||,mathescape=true, fontsize=\small]{nasm}
MLD(R0, ADD1, 1) //load weight |$w_{ij}$|
MLD(R1, ADD2, 0) //load state |$v_{i}$|
ADD(R1, R0, R1) //|$v_{i}=v_{i}+w_{ij}$|
MST(ADD2, R1, 1) //store R1 in |$v_{i}$|
\end{minted}
\vspace{-5pt}
\small Micro-kernel 2: \textit{\textbf{Spike Generation}. See Eq. (\ref{ifneuron}).}
\begin{minted}[frame=single,escapeinside=||,mathescape=true,fontsize=\small]{nasm}
MLD(R0, ADD1, 0) //load state |$v_{i}$|
GTH(R2, R0, R1) //generate spike |$H(v_{i}-$|R1|$)$|
MUL(R3, R2, R0) //|$v_i\times s_{out,i}$|
SUB(R0, R0, R3) //reset state if spike
MST(ADD1, R0, 1) //store R0 in |$v_{i}$|
EVC(R2) //capture event
\end{minted}
\vspace{-5pt}
\caption{Integrate and Fire Neuron}
\label{microifneuron}
\end{listing}

\begin{listing}[t]
\small Micro-kernel 1: \textit{\textbf{Sigma Integration}. See Eq. (\ref{sigma}).}
\begin{minted}[frame=single,escapeinside=||,mathescape=true,fontsize=\small]{nasm}
MLD(R0, ADD1, 1) //load weight |$w_{ij}$|
MLD(R1, ADD2, 0) //load sigma state |$z_{i}$|
MUL(R3, R0, R2) //|$w_{ij}\times o_{in, j}$|, R2|$\xleftarrow{} o_{in, j}$|
ADD(R1, R1, R3) //|$z_{i}=z_{i}+w_{ij}\times o_{in, j}$|
MST(ADD2, R1, 1) //store R1 in |$z_{i}$|
\end{minted}
\vspace{-5pt}
\small Micro-kernel 2: \textit{\textbf{Delta Difference}. See Eq. (\ref{delta}).}
\begin{minted}[frame=single,escapeinside=||,mathescape=true,fontsize=\small]{nasm}
MLD(R0, ADD1, 1) //load sigma state |$z_{i}$|
MLD(R1, ADD2, 0) //load quantize |$f(z_{i}[k-1])$|
MAX(R0, R0, R2) //ReLU |$f(z_{i}[k])=max(z_{i}, 0)$|,R2|$\xleftarrow{} 0$|
DIV(R0, R0, R3) //|$f(z_{i}[k])/q$|, R3|$\xleftarrow{} q$|
RND(R0, R0) //round to closest integer
MUL(R0, R0, R3) //rescale quantization
SUB(R3, R0, R1) //delta |$f(z_i[k]) - f(z_i[k-1])$|
MST(ADD2, R0, 1) //store R0 in quantize |$f(z_{i}[k])$|
EVC(R3) //capture event
\end{minted}
\vspace{-5pt}
\caption{Sigma Delta Neuron}
\label{microsdneuron}
\vspace{-10pt}
\end{listing}

\subsection{Hebbian Learning}
\label{hebbiansec}

Hebbian learning and its variants are bio-inspired unsupervised learning rules that have been extensively used to train shallow SNNs \cite{9892362}. In contrast to backprop-based learning, Hebbian learning schemes do not suffer from update locking and weight transport problems \cite{frenkel202028}, making them better suited for low-complexity on-chip learning \cite{9847074}. Given a layer of spiking neurons with fully-connected connections, the Hebbian learning rule modifies the weight as follows \cite{9180808}:
\begin{equation}
\small
    w_{ij}[k] \xleftarrow{} w_{ij}[k-1] + \eta \times \text{trace}\{s_{out,i} \}[k] \times \text{trace}\{ s_{in,j} \}[k]
    \label{hebbian}
\end{equation}
where $\eta$ is the learning rate and $\text{trace}\{.\}$ is an estimator of the local spiking rate via low-pass filtering:
\begin{equation}
\small
    \text{trace}\{s\}[k] \xleftarrow{} \beta \times \text{trace}\{s\}[k-1] + (1-\beta) \times s[k]
    \label{trace}
\end{equation}
where $\beta$ is the decay constant. Micro-kernels in Components \ref{microhebbian} update the SNN weights at the end of each time step.

% Micro-kernel \ref{microhebbian} and \ref{microtrace} provide the SENeCA micro-codes for Hebbian learning (\ref{hebbian}) and \textit{trace} computation (\ref{trace}).

%Component \ref{microhebbian} provides the SENeCA micro-kernels. Both micro-kernels update per synapse at the end of the time step.

% \begin{listing}[h]
% \begin{minted}[frame=single,escapeinside=||,mathescape=true]{nasm}
% NCP_MLD(R0, ADD1, 0) //load weight |$w_{ij}$|
% NCP_MLD(R1, ADD2, 1) //load |$\text{trace}\{ s_{in,j} \}$| 
% NCP_MUL(R3, R1, R2) //|$\text{trace}\{s_{out,i} \} \times \text{trace}\{ s_{in,j} \}$| 
% NCP_MUL(R5, R3, R4) //store |$\eta \times $|R3 in R5
% NCP_ADD(R6, R0, R5) //store sum in R6 
% NCP_MST(ADD1, R6, 1) //store R6 in |$w_{ij}$|
% \end{minted}
% \vspace{-5pt}
% \caption{\textit{\textbf{Hebbian Learning.} See Eq. (\ref{hebbian}).}}
% \label{microhebbian}
% \end{listing}

% \begin{listing}[h]
% \begin{minted}[frame=single,escapeinside=||,mathescape=true]{nasm}
% NCP_MLD(R0, ADD1, 0) //load |$\text{trace}$| 
% NCP_MLD(R1, ADD2, 1) //load input |$s$|
% NCP_MUL(R4, R2, R0) //|$\beta \times \text{trace}$| 
% NCP_MUL(R5, R3, R1) //|$(1-\beta) \times s$| 
% NCP_ADD(R0, R4, R5) //store sum in R0
% NCP_MST(ADD1, R0, 1) //store R0 in |$\text{trace}$|
% \end{minted}
% \vspace{-5pt}
% \caption{\textit{\textbf{Spike Trace.} See Eq. (\ref{trace}).}}
% \label{microtrace}
% \end{listing}

\begin{listing}[t]
\vspace{5.2pt}
\small Micro-kernel 1: \textit{\textbf{Weight Update.} See Eq. (\ref{hebbian}).}
\begin{minted}[frame=single,escapeinside=||,mathescape=true,fontsize=\small]{nasm}
MLD(R0, ADD1, 0) //load weight |$w_{ij}$|
MLD(R1, ADD2, 1) //load |$\text{trace}\{ s_{in,j} \}$| 
MUL(R1, R1, R2) //|$\text{trace}\{s_{out,i} \} \times \text{trace}\{ s_{in,j} \}$| 
MUL(R1, R1, R3) //|$\eta \times $|R1, R3|$\xleftarrow{} \eta$|
ADD(R0, R0, R1) //update weight 
MST(ADD1, R0, 1) //store R0 in |$w_{ij}$|
\end{minted}
\vspace{-5pt}
\small Micro-kernel 2: \textit{\textbf{Spike Trace Update.} See Eq. (\ref{trace}).}
\begin{minted}[frame=single,escapeinside=||,mathescape=true,fontsize=\small]{nasm}
MLD(R0, ADD1, 0) //load |$\text{trace}$| 
MLD(R1, ADD2, 1) //load input |$s$|
MUL(R0, R0, R2) //|$\beta \times \text{trace}$|, R2|$\xleftarrow{} \beta$|
MUL(R1, R1, R3) //|$(1-\beta) \times s$|, R3|$\xleftarrow{} (1-\beta$)|
ADD(R0, R0, R1) //update |$\text{trace}$|
MST(ADD1, R0, 1) //store R0 in |$\text{trace}$|
\end{minted}
\vspace{-5pt}
\caption{Hebbian Learning}
\label{microhebbian}
\end{listing}

\begin{listing}[t]
\small Micro-kernel 1: \textit{\textbf{Eligibility Trace Update.} See Eq. (\ref{eligibility}).}
\begin{minted}[frame=single,escapeinside=||,mathescape=true,fontsize=\small]{nasm}
MLD(R0, ADD1, 0) //load |$e_{ij}$| 
MLD(R1, ADD2, 1) //load |$\text{trace}\{ s_{in,j} \}$|
MLD(R2, ADD3, 1) //load state |$v_{ij}$|
SUB(R2, R2, R3) //|$v_{ij}-v_{th}$|
ABS(R2, R2) //absolute of R2
GTH(R2, R4, R2) //check R2|$<$|R4, R4|$\xleftarrow{} a_1/2$|
MUL(R2, R2, R5) //R2|$\xleftarrow{} h(v_i)$|, R5|$\xleftarrow{} 1/a_1$|
MUL(R2, R2, R1) //|$h(v_i)\times \text{trace}\{ s_{in,j} \}$|
ADD(R0, R0, R2) //update |$e_{ij}$|
MST(ADD1, R0, 1) //store R0 in |$e_{ij}$|
\end{minted}
\vspace{-5pt}
\small Micro-kernel 2: \textit{\textbf{Weight Update.} See Eq. (\ref{grad-weight}).}
\begin{minted}[frame=single,escapeinside=||,mathescape=true,fontsize=\small]{nasm}
MLD(R0, ADD1, 0) //load weight |$w_{ij}$| 
MLD(R1, ADD2, 1) //load |$e_{ij}$|
MLD(R2, ADD3, 1) //load feedback error
MUL(R1, R3, R1) //|$\eta \times e_{ij}$|, R3|$\xleftarrow{} \eta$|
MUL(R2, R2, R1) //|$\eta\times  e_{ij}\times \Sigma_k b_{ik}\times y_k$|
SUB(R0, R0, R2) //update weight
MST(ADD1, R0, 1) //store R0 in |$w_{ij}$|
\end{minted}
\vspace{-5pt}
\caption{Gradient-based Online Learning (e-prop)}
\label{microgradient}
\vspace{-10pt}
\end{listing}

% Need to update this section with current implementation
\subsection{Gradient-based Online Learning (e-prop)}

Gradient-based online learning performs end-to-end learning in SNN by estimating gradients using only local information~\cite{bellec2020solution,tang2021biograd, bohnstingl2022online,neftci2017event}. Here, we show the e-prop learning~\cite{bellec2020solution} in SENeCA as an example. First, the eligibility trace $e_{ij}$ combines pre- and post-synaptic activities:
\begin{equation}
\small
    e_{ij}[k] \xleftarrow{} e_{ij}[k-1] + h(v_i[k])\times \text{trace}\{ s_{in,j} \}[k]
\label{eligibility}
\end{equation}
where $h$ is the surrogate gradient function. The weight updates when there are error events from the supervised signal:
\begin{equation}
\small
    \triangle w_{ij} = -\eta\times  e_{ij}\times \Sigma_k b_{ik}\times y_k
\label{grad-weight}
\end{equation}
where $b_{ik}$ is the feedback weight and $y_k$ is the error events from the output layer. We implemented the learning rule using four SENeCA micro-kernels, with the first micro-kernel in Component \ref{microgradient} updating every time step using a rectangular function for $h$ as introduced in~\cite{wu2018spatio}, and the second micro-kernel in Component \ref{microgradient} updates when there is a supervised signal available. Additionally, we use micro-kernel 2 in Component \ref{microhebbian} to compute $\text{trace}\{ s_{in,j} \}$ and micro-kernel 1 in Component \ref{microsdneuron} to compute $\Sigma_k b_{ik}\times y_k$.

\subsection{Efficient Synaptic Operation with Hierarchical Memory}
% Need to rewrite this section
\label{intweight}
The measurement results show memory accesses dominate the total energy consumption for neuron processing. Hierarchical memory architecture in SENeCA allows for data-reuse in NPE register-files and therefore reducing more expensive SRAM accesses. This reduction is achieved using quantized weights and multi-event processing. Using quantized weights (4-bit or 8-bit) reduces the number of SRAM reads per weight. However, there is an overhead as the weight needs to be converted into BF16 using the I2F instruction for computation. As another example of data reuse in the NPEs, processing multiple events in one iteration also reduces the SRAM accesses. The neuron state becomes stationary on the NPEs, avoiding 
 frequently accessing the states from the SRAM. Using fully integer operations on INT4 weights and INT8 states further reduce memory accesses, and thereby decrease energy cost.

Table \ref{tableweight} shows the average energy cost per IF neuron synaptic operation (i.e., spike integration) when using the integer weights with one and four event processing. By exploiting hierarchical memory, SNN algorithms in SENeCA can potientially achieve lower energy costs compared to existing digital neuromorphic processors without hierarchical memory~\cite{merolla2014million,davies2018loihi,moreira2020neuronflow} (see Table \ref{tableweight} bottom row).

% \begin{table}[t]
% \vspace{5.2pt}
% \small
%   \caption{Neuromorphic Components Energy Consumption}
%   \label{tablecomponents}
%   \centering
%   \begin{tabular}{|l|c|c|c|}
%     \hline
%     Component & Micro-kernel & Energy (pJ) & Frequency \\
%     \hline
%     \multirow{2}{5em}{IF Neuron} & 1 & 12.1 & event\\
%     & 2 & 11.05 & time step\\
%     \hline
%     \multirow{2}{5em}{SD Neuron} & 1 & 13.4 & event\\
%     & 2 & 16.72 & flexible\\
%     \hline
%     \multirow{2}{5em}{Hebbian Learning} & 1 & 14.7 & time step\\
%     & 2 & 14.7 & time step\\
%     \hline
%     \multirow{2}{5em}{Gradient Learning} & 1 & 21.38 & time step\\
%     & 2 & 18.3 & flexible\\
%     \hline
%   \end{tabular}
% \end{table}

\begin{table}[t]
\vspace{5.2pt}
\small
  \caption{Neuromorphic Components Energy Consumption}
  \label{tablecomponents}
  \centering
  \begin{tabular}{|l|c|c|c|}
    \hline
    Component & Micro-kernel & Energy (pJ) & Frequency \\
    \hline
    \multirow{2}{5em}{IF Neuron} & 1 & 12.7 & event\\
    & 2 & 13.2 & time step\\
    \hline
    \multirow{2}{5em}{SD Neuron} & 1 & 14.1 & event\\
    & 2 & 19.7 & flexible\\
    \hline
    \multirow{2}{5em}{Hebbian Learning} & 1 & 15.5 & time step\\
    & 2 & 15.5 & time step\\
    \hline
    \multirow{2}{5em}{Gradient Learning} & 1 & 22.9 & time step\\
    & 2 & 19.2 & flexible\\
    \hline
  \end{tabular}
\end{table}

% \begin{table}[t]
% \small
%   \caption{Energy per Synaptic Operation in SENeCA}
%   \label{tableweight}
%   \centering
%   \begin{tabular}{|l|c|c|c|c|c|}
%     \hline
%    \hspace{-8.5pt} Weight, \textit{1 Event}  &  BF16 & Int8 & Int4\\
%     \hline
%     Energy (pJ) & 12.1 & 10.92 & 10.02\\
%     \hline
%     \hline
%     \hspace{-8.5pt} Weight, \textit{4 Events} &  BF16 & Int8 & Int4\\
%     \hline
%     Energy (pJ) & 6.7 & 5.52 & 4.62\\
%     \hline
%     \hline
%     \textbf{Hardware} &  Loihi \cite{davies2018loihi} & TrueNorth \cite{merolla2014million} & NeuronFlow \cite{moreira2020neuronflow}\\
%     \hline
%     Energy (pJ)\tablefootnote{Results without synaptic operation details, may use different neuron model.} & 23 & 2.5 & 20\\
%     \hline
%   \end{tabular}
% \end{table}

\begin{table}[t]
\small
  \caption{Energy per Synaptic Operation in SENeCA}
  \label{tableweight}
  \centering
  \begin{tabular}{|l|c|c|c|c|}
    \hline
   \hspace{-8.5pt} Weight, \textit{1 Event}  &  BF16 & Int8 & Int4 & Int4\tablefootnote{Using INT8 state and integer arithmetic operations.}\\
    \hline
    Energy (pJ) & 12.7 & 11.95 & 11.03 & 5.63\\
    \hline
    \hline
    \hspace{-8.5pt} Weight, \textit{4 Events} &  BF16 & Int8 & Int4 & Int4$^2$\\
    \hline
    Energy (pJ) & 7.0 & 6.25 & 5.33 & 2.78\\
    \hline
    \hline
    \textbf{Hardware} &  Loihi \cite{davies2018loihi} & TrueNorth \cite{merolla2014million} & \multicolumn{2}{c|}{NeuronFlow \cite{moreira2020neuronflow}}\\
    \hline
    Energy (pJ)\tablefootnote{Results without synaptic operation details, may use different neuron model.} & 23 & 2.5 & \multicolumn{2}{c|}{20} \\
    \hline
  \end{tabular}
\end{table}

\section{Application-Level Benchmarking}
\label{applevel}
To illustrate how the cost of neuromorphic components reported in Table \ref{tablecomponents} can be used to reliably estimate the energy cost of solving downstream tasks and optimize the algorithm based on application needs, we perform an application-level benchmarking of SNN algorithms for video processing and online Hebbian learning.

\subsection{Sigma Delta Network for Video Processing}

Sigma Delta networks can result in more than 90\% synaptic operation sparsity when performing video-based human action recognition without sacrificing accuracy~\cite{yousefzadeh2022delta}. Here, we compute the energy cost of employing SD neurons in ResNet-50~\cite{he2016deep} and MobileNet~\cite{howard2017mobilenets} on SENeCA, for efficient video processing using the UCF-101 human action recognition dataset~\cite{soomro2012ucf101}. Using Table \ref{tablecomponents}, we calculate the average energy cost of the networks per frame, as shown in Table \ref{tablesdnetwork}, by counting the number of synaptic operations (sigma) and neuron output evaluations (delta). Compared to the estimated energy cost in~\cite{yousefzadeh2022delta}, the precise instruction-level results given here reflect in a more accurate way the actual energy cost of hardware processing. Although a single delta operation requires more instructions than sigma, Table \ref{tablesdnetwork} shows that sigma operations cost much more energy compared to delta, due to the difference in execution dimensionality. Therefore, an algorithm designer can reduce the number of events using a complex delta unit with only negligible energy overheads.

% \begin{table}[t]
% \vspace{5.2pt}
% \small
%   \caption{Energy consumption per frame for SD networks}
%   \label{tablesdnetwork}
%   \centering
%   \begin{tabular}{|l|c|c|c|}
%     \hline
%     Network Model &  Sigma Energy ($\mu$J) & Delta Energy ($\mu$J)\\
%     \hline
%     ResNet-50 & 3330.1 & 153.8\\
%     \hline
%     MobileNet & 1703.3 & 88.6\\
%     \hline
%   \end{tabular}
% \end{table}

\begin{table}[t]
\vspace{5.2pt}
\small
  \caption{Energy consumption per frame for SD networks}
  \label{tablesdnetwork}
  \centering
  \begin{tabular}{|l|c|c|c|}
    \hline
    Network Model &  Sigma Energy ($\mu$J) & Delta Energy ($\mu$J)\\
    \hline
    ResNet-50 & 3504.0 & 181.2\\
    \hline
    MobileNet & 1792.3 & 104.4\\
    \hline
  \end{tabular}
\end{table}

\begin{figure}
\centering
\includegraphics[scale=0.4]{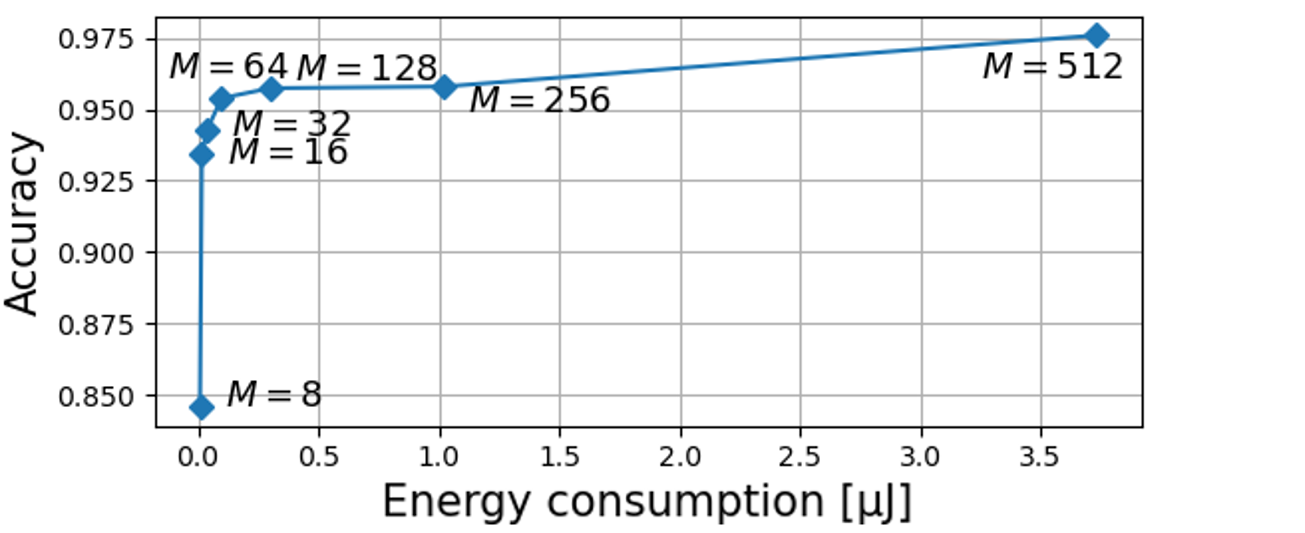}
\caption{Accuracy vs. energy consumption of SNN-Hebbian learning execution (one time-step), with $M$ output neurons. %The higher the number of output neuron $M$, the higher the accuracy but the higher the energy consumption.
}
\label{fig:tradeof}
\end{figure}

\subsection{Unsupervised Hebbian Learning for Digit Classification}
For demonstrating the cost of online learning, we consider a canonical digit classification task \cite{620583} with a dataset composed of $1797$ instances of $8\times 8$ grayscale images normalized between $0$ and $1$ \cite{pedregosa2011scikit}. We flatten each image to a $64$-dimension grayscale vector and encode each entry of the vector into a 100 time-step spike train using Poisson spike encoding.

In SENeCA, we implement a modified version of the SNN architecture proposed in \cite{9892362}, where we use the IF neuron model of Section \ref{IFneuron} and the Hebbian learning model of Section \ref{hebbiansec} to replace the leaky IF neurons and the Hebbian-like STDP rule used in \cite{9892362}. We can then estimate the energy consumption of the network in function of the number of output neurons $M$ and the input dimension $N$, by counting the number of instructions executed for each completed SNN execution step (i.e., forward propagation of the input spikes, recurrent propagation of the output spikes, feedback propagation of the output spikes and all Hebbian learning mechanisms in \cite{9892362}). Then, energy consumption is found by relating the number of occurrences of each instruction with the energy measurements provided in Table \ref{tableinst} and \ref{tablecomponents}. In the illustrative case of the canonical digit classification dataset \cite{620583}, the input data dimension is $N=64$ and the output dimension $M$ can be arbitrarily chosen, leading to a trade-off between energy consumption and classification accuracy, as shown in Figure \ref{fig:tradeof}. The ability to accurately generate this trade-off gives algorithm designers the opportunity to \textit{optimize} the network size based on the need of the application \textit{without} the overhead of hardware deployment.

% Doing so, the following formula for energy estimation is obtained:
% \begin{multline}
%   E_{tot} = \frac{M}{8} (N+2M+1) \times E_{ca} + \frac{M}{8} M \times E_{cpy}
%   \\ + (\frac{M}{8} + \frac{2N}{8}) \times E_{pp} + (\frac{3M}{8} + \frac{5N}{8}) \times E_{psc} + (\frac{2M}{8} + \frac{2N}{8}) \times E_{cv}
%     \\ \hspace{-10pt} + (\frac{2M}{8} + \frac{2N}{8}) \times E_{st} + (\frac{M}{8}(N+M) + \frac{N}{8}M) \times E_{hebb}
%     \\ + \frac{M}{8} 2M \times E_{ld} + (\frac{M}{8} (3M+2N+3) + \frac{NM}{8}) \times E_{sca} 
% \end{multline}
% where the divisions by $8$ come from the fact that the neural co-processor in SENeCA can update $8$ floating point numbers simultaneously.

\section{Conclusion}
\label{conc}
This paper presents a practical approach to properly benchmark SNN algorithms using the neuron processing instruction set of the SENeCA neuromorphic architecture. We strongly believe that the instructions and micro-kernels provided here, together with their precise energy measurements, will allow a reliable estimate of energy consumption at algorithm design time. We hope that this work will greatly help algorithm designers to conveniently benchmark the hardware costs of their various SNN algorithms, and will enable further optimization of these costs via effective algorithm-hardware co-design.

\section*{\small{Acknowledgment}}
\small{This work is partially funded by research and innovation projects ANDANTE (ECSEL JU under grant agreement No876925), DAIS (KDT JU under grant agreement No101007273) and MemScale (Horizon EU under grant agreement 871371). The JU receives support from the European Union’s Horizon 2020 research and innovation programme and Sweden, Spain, Portugal, Belgium, Germany, Slovenia, Czech Republic, Netherlands, Denmark, Norway and Turkey.}

%TBD

\bibliographystyle{IEEEtran}
\bibliography{ref}

\end{document}